\documentclass[a4paper,fleqn]{cas-dc}

\usepackage[numbers]{natbib}
\usepackage{float}
\usepackage{graphicx}
\usepackage{caption}
\usepackage{times}
\usepackage{xcolor}
\usepackage{color}
\usepackage[switch]{lineno}

\usepackage{algorithm,algpseudocode,algorithmicx}
\usepackage{subfigure}
\usepackage{booktabs}
\usepackage{pgfgantt}
\usepackage{amsfonts}
\usepackage{amsmath}
\usepackage[justification=centering]{caption}
\usepackage{makeidx}
\makeindex

\begin{document}
% \linenumbers
\let\WriteBookmarks\relax
\def\floatpagepagefraction{1}
\def\textpagefraction{.001}
\shorttitle{DRAFT}
\shortauthors{Jia Liu et~al.}

\title [mode = title]{Multi-objective Search of Robust Neural Architectures against Multiple Types of Adversarial Attacks}                    
\author[1]{\color{black}Jia Liu}
%\ead{xuqiaoyichn@163.com}

\author[1]{\color{black}Yaochu Jin}
\cormark[1]

%
%\credit{Conceptualization of this study, Methodology, Software}
%
\address[1]{Department of Computer Science, University of Surrey, Guildford, GU2 7XH, United Kingdom}
% \address[2]{xxxxxxx}

\cortext[cor1]{Corresponding author}
% \cortext[0]{Email: xx@xxx}

\begin{abstract}
Many existing deep learning models are vulnerable to adversarial examples that are imperceptible to humans. To address this issue, various methods have been proposed to design network architectures that are robust to one particular type of adversarial attacks. It is practically impossible, however, to predict beforehand which type of attacks a machine learn model may suffer from. To address this challenge, we propose to search for deep neural architectures that are robust to five types of well-known adversarial attacks using a multi-objective evolutionary algorithm. To reduce the computational cost, a normalized error rate of a randomly chosen attack is calculated as the robustness for each newly generated neural architecture at each generation. All non-dominated network architectures obtained by the proposed method are then fully trained against randomly chosen adversarial attacks and tested on two widely used datasets. Our experimental results demonstrate the superiority of optimized neural architectures found by the proposed approach over state-of-the-art networks that are widely used in the literature in terms of the classification accuracy under different adversarial attacks.
\end{abstract}

%\includegraphics[width=16cm,keepaspectratio]{xxx}

%\item Constructing a gas leakage dataset by CFD simulation 
%\item Extension and implementation of ANN-based STE for obstacle cases
%\item High performance one-step CNN-based STE method
%\item Comprehensive experiments for verification


\maketitle
\section{Introduction}
Deep neural networks (DNNs) have been successfully used to deal with various computer vision tasks, such as image recognition, object detection and segmentation. However, it is recognized that neural networks are vulnerable to adversarial attacks. For example, some visually imperceptible perturbations to the images generated by the fast gradient sign method (FGSM) \cite{goodfellow2014explaining}
%, basic iterative method (BIM) \cite{kurakin2016adversarial}, C\&W attack \cite{carlini2017towards}, DeepFool \cite{moosavi2016deepfool} 
can completely mislead the classifiers. Such vulnerability must be fixed before deep learning models can be adopted in safety-critical applications.
% Recently, significant research efforts have been devoted to the design of neural networks that can defend against adversarial attacks. 
Adversarial training \cite{goodfellow2014explaining} is one main counter-measure, during which adversarial examples are generated in every step of training and then are injected into the training dataset. 
% Adversarial training has shown to be able to improve both the robustness and the accuracy of DNNs to defend a known type of attacks.
Nevertheless, adversarial training with FGSM has found to be ineffective when the adversarial attacks are iterative, such as the basic iterative method (BIM) \cite{kurakin2016adversarial}.
% new adversarial examples can mislead the networks so that adversarial training is usually considered a kind of regularization method to avoid overfitting. 
Papernot et al. \cite{papernot2016distillation} adapted defensive distillation to address the DNN's vulnerability to adversarial perturbations. 
% Firstly, an initial network is trained with a temperature $T$ using standard techniques to produce soft labels. Then the soft labels which include additional knowledge about the classes compared to a class label, are used to train a distilled network at a temperature $T$ on the same data $X$. 
Their method can make DNNs robust to some adversarial attacks but was shown later on to be ineffective against optimization-based attacks such as C\&W attacks \cite{carlini2016defensive}. 
% Besides, detection methods such as feature squeezing \cite{xu2017feature} and example reforming \cite{meng2017magnet} can also be used to identify adversarial examples. 
Several methods focus on using an auxiliary tool before feeding images to the classifier, including JPEG compression \cite{dziugaite2016study}, feature squeezing \cite{xu2017feature}, defense-GAN \cite{samangouei2018defense}, and autoencoder-based denoising \cite{liao2018defense}. However, these defense techniques lead to shattered gradients or obfuscated gradients \cite{athalye2018obfuscated}, which can be evaded by adaptive attacks. The research on adversarial robustness has been facing with an ``arms race'' between defenders and attackers, i.e., a defense method against a certain attack was soon evaded by new attacks and vice versa. 

Little research has been reported on tackling with adversarial robustness of neural networks from an architectural design perspective.  In \cite{vargas2019evolving}, robust architecture search (RAS) was proposed to search the architectures that are robust to transferable black-box attacks. The fitness of the models discovered by RAS is evaluated as the validation accuracy on clean images plus the attack resilience on 2812 adversarial examples. The evolved architecture was shown to have achieved robust performance without any defense techniques. However, only black-box attacks are used to evaluate the robustness of architectures. Based on the one-shot NAS \cite{cai2019once}, Guo et al. \cite{guo2020meets} focused on improving the network robustness during NAS by employing the projected gradient descent (PGD) method with seven steps of adversarial training.
A family of robust architectures (RobNets) that are discovered by this approach are more resistant to white-box and black-box attacks than some handcrafted models. However, only three operations are used in RobNets to lift the burden of one-shot NAS-based adversarial training. In \cite{devaguptapu2020adversarial}, Devaguptapu et al. analyzed the adversarial robustness of both hand-crafted and NAS-based architectures. However, their models are trained without any defense techniques and consequently none of the models can perform well under strong adversarial attacks.  

To address the above limitations, this work aims to improve the robustness of DNNs by designing new architectures. A multi-objective evolutionary algorithm taking both prediction accuracy and robustness into account is employed to search for neural network architectures that are less sensitive to multiple types of adversarial attacks. The core contributions of this research can be summarized as follows:
\begin{itemize}
\item We design a new measure to evaluate the robustness of neural architectures, which can evaluate the performance under various adversarial attacks including four white-box attacks and a transferable black-box attack. The proposed measure uses the overall robustness of the 18 popular hand-crafted networks on five adversarial attacks as the baseline to quantify the robustness against multiple types of adversarial attacks.
% \item We use a transfer-based black-box attack crafted from FGSM on VGG-16 to evaluate the robustness of an architecture against black-box attack.
\item The proposed robustness measure against multiple adversarial attacks is adopted as an objective function in addition to classification performance on clean data to guide the multi-objective evolutionary algorithm to search for robust architectures.
\item Extensive empirical studies are performed on CIFAR-10 and CIFAR-100 datasets to demonstrate the effectiveness of the proposed method.
%\item The NSGA-II is used to target the success rate of white-box attacks or (/ and) black-box attacks to guide the search for a more robust structure.
%\item Explore the extent to which different network operations are resistant to attack, discover key operations, and guide future selected search Spaces.
\end{itemize}

The rest of this paper is organized as follows. The next section will briefly introduce the related techniques used in this work. In Section 3, we detail the proposed approach to multi-objective robust neural architecture search, including a performance measure for assessment of robustness under multiple attacks. Experimental settings are presented in Section 4, followed by a description of the experimental results in Section 5. Finally, we draw conclusions and outline future work in Section 6.

\section{Background}
This section reviews the background of the present work, including adversarial attacks, adversarial training and multi-objective neural architecture search. We begin with an introduction to adversarial example crafting techniques related to both white-box and black-box attacks, which will be used in this work.

\subsection{Adversarial Attacks}
Adversarial attacks refer to methods in which adversaries deliberately craft imperceptible adversarial examples to mislead the classifier. Adversarial attacks can be grouped into two categories white-box attacks and black-box attacks, depending on whether information about the classifier is used. In the following, we focus our discussion on adversarial attacks for image classification.
\subsubsection{White-box Attacks}\label{watk}
\begin{figure*}[t]
\begin{center}
\subfigure[original]{
\includegraphics[width=.16\textwidth]{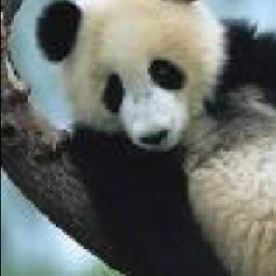}
\label{a}
%\caption{fig1}
}
\quad
\subfigure[FGSM.]{
\includegraphics[width=.16\textwidth]{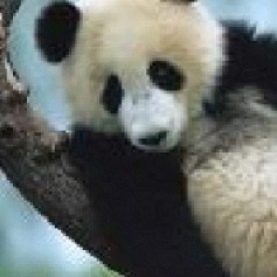}
\label{b}
}
\quad
\subfigure[BIM]{
\includegraphics[width=.16\textwidth]{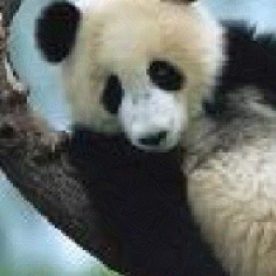}
\label{c}
}
\quad
\subfigure[PGD]{
\includegraphics[width=.16\textwidth]{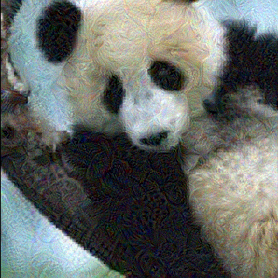}
\label{f}
}
\quad
\subfigure[FFGSM]{
\includegraphics[width=.16\textwidth]{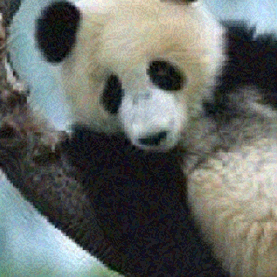}
\label{g}
}
\caption{Examples of adversarial images generated by different adversarial attacks for Inception V3 \cite{szegedy2016rethinking}. (a) The original clean image in ImageNet: classified as ``giant panda''; (b) FGSM: classified as ``giant panda''; (c) BIM: classified as ``standard poodle''; (d) PGD: classified as ``toy poodle''; (e) FFGSM: classified as ``giant panda'' }
\label{panda}
\end{center}
\end{figure*}
White-box attacks assume that the adversary knows detailed information of the targeted models, including model architecture, hyper-parameters, gradients, and training data. In the following, we use $X^*$ and $X$ denote the adversarial and clean examples, respectively. Then, ${\nabla}_{X}$ measures the gradient of the loss function $l$ with respect to $X$. 
% \begin{itemize}
% \item 

\textbf{Fast Gradient Sign Method (FGSM).}
It is a one-step and non-target attack, which generates adversarial examples by adding perturbations along the direction of the sign of gradient at each pixel  \cite{goodfellow2014explaining}. The generated adversarial examples can be calculated by
\begin{eqnarray}
{X^*} = X + \epsilon \cdot \textrm{sign}({\nabla}_{X}l(X,y_{true}))
\label{2.1}
\end{eqnarray}
where $\epsilon$ is a hyper-parameter that controls the magnitude of the disturbance.
% \item 

\textbf{Basic Iterative Method (BIM).}
This is an iterative version of FGSM (also called I-FGSM), in each step of which the pixel values are clipped to a certain range  \cite{kurakin2016adversarial}:
\begin{eqnarray}
	{X^*_0} &=& X,\\
	{X^*_{n+1}} &=& Clip_{X,\epsilon}\lbrace{X^*_{n} + \alpha \cdot \textrm{sign}	({\nabla}_{X}l(X^*_{n},y_{true}))}\rbrace
\label{2.2}
\end{eqnarray}
where $Clip_{X,\epsilon}(A)$ denotes element-wise clipping $A$, with $A_{i,j}$ clipped to the range $[X_{i,j}-\epsilon, X_{i,j}+\epsilon]$,  $\alpha$ denotes the step size, and ${X^*_{n+1}}$ denotes the adversarial example after $n$-steps.

% \item 
\textbf{Projected Gradient Descent (PGD).} The PGD attack \cite{madry2018towards}, which combines randomized initialization with multi-step attacks, is one of the strongest adversarial attack against adversarial training. The adversarial examples generated by the PGD attack can be expressed as:
\begin{eqnarray}
{X^*_{0}} &=& X + \mathcal{U}(-\epsilon, \epsilon), \\
{X^*_{n+1}} &=& \Pi_{X,\epsilon}\lbrace{X^*_{n} + \alpha \cdot \textrm{sign}({\nabla}_{X^*_{n}}l(X^*_{n},y_{true}))}\rbrace
\label{2.3}
\end{eqnarray}
where $\mathcal{U}$ is a uniform distribution and $\Pi_{X,\epsilon}(B)$ refers the projection to $B(X, \epsilon)$.
% \item 

\textbf{FFGSM.}
The FFGSM attack \cite{wong2020fast} is FGSM starting from a random noise and its step size $\alpha$ is usually larger than $\epsilon$.
\begin{eqnarray}
{X^*_{0}} &=& X + \mathcal{U}(-\epsilon, \epsilon), \\
{X^*} &=& \Pi_{X,\epsilon}\lbrace{X^*_{0} + \alpha \cdot \textrm{sign}({\nabla}_{X^*_{0}}l(X^*_{0},y_{true}))}\rbrace
\label{2.4}
\end{eqnarray}
% \end{itemize}

Fig. \ref{panda} illustrates the adversarial examples generated using the above mentioned methods. After adding the perturbations, the images are visually similar and human can recognize that it is a giant panda. But attacks, especially the strong iterative adversarial attacks like PGD, can mislead the model to make a wrong classification. We employ the above-mentioned attacks since most of them are fast and effective. In this work, we do not use C\&W attack \cite{carlini2016defensive} because it is so strong that the accuracy of nearly all neural architectures will become 0. In addition, the C\&W attack is computationally very intensive, making it less suited for guiding neural architecture search.
\subsubsection{Black-box Attacks}\label{batk}
In this case, it is assumed that the adversary only knows the outputs of the model to perform a black-box attack. The adversaries keep feeding samples into the model and observe the output to find the relationship between the input and the output. Compared to white-box attacks, black-box attacks are more practical in real scenarios because adversaries usually do not know the exact model information. White-box attacks have been found to be transferable to attack black-box models \cite{papernot2016transferability}. Such transfer-based attacks do not rely on model information but need information about the training data. This data is used to train a fully observable substitute
model, from which adversarial perturbations can be synthesized. For example, VGG-16 and VGG-19 models achieve significantly better transferability than other models for all attacking methods \cite{su2018robustness}. VGG models are thus a good starting point for mounting transfer-based black-box attacks.

\subsection{Adversarial Training}
Adversarial training (AT) \cite{goodfellow2014explaining} is a widely used technique to improve the adversarial robustness of deep learning models. The basic idea of AT is to create and incorporate adversarial samples during the training phase. It was first used with a gradient-based single-step adversarial attack (FGSM) \cite{goodfellow2014explaining}. Later, the models trained with FGSM-based adversarial training (FGSM-AT) was found to tend to overfit and remain vulnerable to stronger attacks such as iterative adversarial attacks (like BIM and PGD). More recently, PGD-based adversarial training (PGD-AT) \cite{madry2018towards} has shown to be able to provide strong adversarial robustness and become popular for defending strong adversarial attacks. Despite the effectiveness of PGD-AT, a critical downside of PGD-AT is that it is time-consuming \cite{shafahi2019adversarial}. In addition to the gradient computation needed to update the network parameters, each stochastic gradient descent (SGD) iteration requires multiple gradients computations to produce adversarial images. For example, the PGD-AT on CIFAR-10 dataset takes about 80 hours. Furthermore, a recent study \cite{rice2020overfitting} conducted extensive experiments on adversarially trained models and demonstrated that the performance gain from almost all recently proposed algorithmic modifications to PGD adversarial training is not better than a simple piecewise learning rate schedule and early stopping to prevent overfitting.

To speed up adversarial training and prevent catastrophic overfitting, Wong et al. \cite{wong2020fast} proposed fast adversarial training (FastAdv) that combines FGSM-AT with random initialization and its effectiveness is similar to a PGD-based training. They also employ cyclic learning rates \cite{smith2019super} to reduce the total number of epochs needed for convergence and further speed up computations. Although the modification is simple, the underlying reason for its success remains unclear. To further understand FastAdv, Li et al. \cite{li2020towards} conduct experiments to show that the key to the success of FastAdv does not lie in the avoidance of catastrophic overfitting, but the retaining of the robustness when catastrophic overfitting occurs. Then a simple remedy of FastAdv was proposed, named FastAdv+, making it possible to train it for a large number of epochs without sacrificing efficiency. 
They also proposed FastAdvW, which revisits a previously developed technique, FGSM-AT as a warmup \cite{wang2019convergence}, and combines it with their training strategy to further improve performance with a small additional computational overhead. The resulting method obtains better performance compared to the state-of-the-art approach, PGD-AT \cite{madry2018towards}, while consuming much less training time.

\subsection{Multi-objective Neural Architecture Search}
Neural architecture search (NAS) aims to automatically discover high-performing network architectures within given search spaces by using an effective search strategy such as reinforcement learning, evolutionary algorithms (EAs), and Bayesian optimization. Compared with designing DNNs manually, NAS methods can search architectures without using much domain knowledge. Actually, the research on automated neural network design can be traced back to 1990s, when EAs were used to evolve the structure of neural networks \cite{Schaffer1992combine}, which is known as neuro-evolution. For instance, the NEAT \cite{stanley2002evolving} and its variants \cite{6792316} evolve network topology along with weights to improve efficiency.
%Due to limited computing resources, the structure of the network is usually fixed, and most researches focus more on hyper-parameter optimization.
More recently, reinforcement learning (RL) was employed to search network architectures \cite{baker2016designing, zoph2016neural}, which, however, is often computationally expensive. To improve the search efficiency, Bayesian optimization frameworks were used for neural architecture search, including NASBOT \cite{kandasamy2018neural}, BANANAS \cite{white2019bananas} and GP-NAS \cite{li2020gp}, which use a Gaussian process or an ensemble meta neural network as a surrogate model to provide predictive uncertainty estimates for architecture performance. Besides, gradient-based methods like DAS \cite{shin2018differentiable}, GNAS \cite{dong2019searching} and DARTS \cite{liu2018darts} were developed to accelerate the search process. However, most of these techniques consider NAS as a single-objective optimization problem, i.e., maximizing the classification accuracy, which do not take other performances into account. 

Most recently, several NAS algorithms have been proposed to investigate the multi-objective network architecture search, which typically employed an EA or RL.
NEMO \cite{kim2017nemo} adopts the NSGA-II framework to handle the trade-off between accuracy and runtime.
LEMONADE \cite{elsken2018efficient} jointly maximizes the predictive performance and minimizes the number of parameters.
MONAS \cite{hsu2018monas} considers classification accuracy and power consumption as the rewards in RL.
MNasNet \cite{tan2019mnasnet} also uses the RL approach to search for mobile models with the best trade-offs between accuracy and latency. They
use a customized weighted product method to approximate the Pareto optimal solutions so that the bi-objective problem is converted into a single objective one.
To optimize FLOPs and accuracy simultaneously, NSGA-II and multi-objective particle swarm optimization have been employed in NAS \cite{zhu2019multi, lu2019nsga, wang2019evolving}.
CARS \cite{yang2020cars} proposes pNSGA-III that simultaneously optimizes the number of parameters, the classification accuracy, and the speed of the accuracy.

Despite the exciting progress, previous work on multi-objective NAS mainly consider model size, latency, complexity and energy consumption. Not much work on multi-objective NAS for robustness and resistance of the architectures against the adversarial attacks has been reported. 

\section{Proposed Method}
DNNs are found vulnerable to adversarial attacks, and a model may have better performance on an attack but worse performance on another. To discover architectures that are less sensitive to multiple types of adversarial attacks, one intuitive approach is to consider the success rate of different attacks as multiple objectives. However, if the robustness to each type of attacks is considered as one objective, the optimization task may become unnecessarily complex. Another more practical approach is to integrate the performance measures for different attacks into one single objective that can represent the overall robustness. Unfortunately, it is already very time-consuming to train a neural architecture to evaluate its performance in the presence of one type of attacks only. Training architectures in the presence of a large number of adversarial attacks will make performance evaluations in NAS  computationally prohibitive.

To address the above challenges, we propose an algorithm for multi-objective search for robust architectures, termed MORAS in short. MORAS is designed to search for neural architectures that are less sensitive to different white-box and black-box adversarial attacks. 
% Fig. \ref{MORAS-W} shows the process of generating adversarial images in MORAS. 
We randomly select one attack among four white-box adversarial attacks and one transferable black-box adversarial attack that generates adversarial examples by attacking VGG-19 using FGSM. 
% All the examples are saved as black-box adversarial examples to attack the target model to reduce the computational budget. 
Then we propose a measure for the robustness performance of a neural architecture in the presence of different types of attacks. Details of the proposed robustness measure will be presented in Section \ref{obj}.

% As discussed in Section \ref{batk}, white-box adversarial attacks are difficult to achieve since an adversary may not have access to know all the information of the target model. Hence, we propose MORAS-B which aims to search architectures that are less vulnerable to black-box adversarial attacks. Fig. \ref{MORAS-B} shows the process of generating adversarial images in MORAS-B. Different from MORAS-W, we attack models using transferred adversarial examples generated by FGSM on VGG-16. The evaluation function $f_{2b}$ will be detailedly presented in subsection \ref{evablk}.

% \begin{figure*}[!t]
% \centering
% %\includegraphics[width=.5\textwidth]{flowchart.png}
% \includegraphics[width=.9\textwidth]{adversarial_crafting}
% \caption{Framework for adversarial crafting in MORAS.}
% \label{MORAS-W}
% \end{figure*}

In the following, we will elaborate on the main components of the proposed algorithm. We start with the encoding strategy, followed by a description of the objective functions, in particular of a robustness measure we propose. Lastly, we will describe the overall framework of the proposed algorithm.

\subsection{Encoding}\label{3.1}
An efficient encoding strategy and a proper search space are of great importance for the performance of NAS. There are three commonly used search spaces in NAS research \cite{kyriakides2020introduction}, i.e., global search space, micro or cell-based search space, and hierarchical search spaces. In the global search space, the number of layers, the type of operation, and hyper-parameters can be explored \cite{xie2017genetic, liang2018evolutionary, cai2018efficient}. Arbitrary networks can be generated in such spaces, and consequently the search space may be very large. To alleviate this problem, numerous recent algorithms \cite{zoph2018learning, liu2018darts} make use of good cells already discovered in the literature instead of using wild basic operators. Similar to the repeated patterns in hand-crafted structures, in a micro or cell-based search space, a NAS algorithm only needs to search for the internal structure of the cell, since the cells connect in a predefined way. Such reduced spaces can simplify the search process and improve the search efficiency. %, and most search strategies can use this cell-based search space because of its flexibility. 
For example, RobNets \cite{guo2020meets} adopt a cell-based search space, in which the candidate operations consist of only $3\times3$ separable convolution, identity, and zero without restricting the maximal number of operations between two intermediate nodes to be one.
The hierarchical search space proposed in \cite{liu2017hierarchical, tan2019mnasnet} aims to keep a balance between the macro search and micro search, while RAS \cite{vargas2019evolving} uses multiple populations (thus hierarchical search spaces) composed of layer population, block population and model population.
 
In this work, we employ the cell-based search space consisting of normal and reduction cells as proposed by Zoph et al. \cite{zoph2018learning}, which has been commonly used in recent years. In this cell-based search space, each cell consists of $n_p$ nodes. The first two nodes are the inputs from the previous cells in the hyper-architecture. Each of the rest nodes contains two edges as inputs, such that each cell forms a connected DAG. Each edge can take one of predefined operations, including $3\times3$ max pooling, $3\times3$ average pooling, identity, $3\times3$, $5 \times5$ and $7\times7$ separable convolutions, and $3\times3$ and $5 \times5$ dilated separable convolutions. Then the normal and reduction cells are stacked together to build the overall architecture. 
% Our proposed methods for evaluating the robustness of architecture can be extended to any multi-objective NAS framework. 

\begin{figure*}[htbp]
\centering
\includegraphics[width=.9\textwidth]{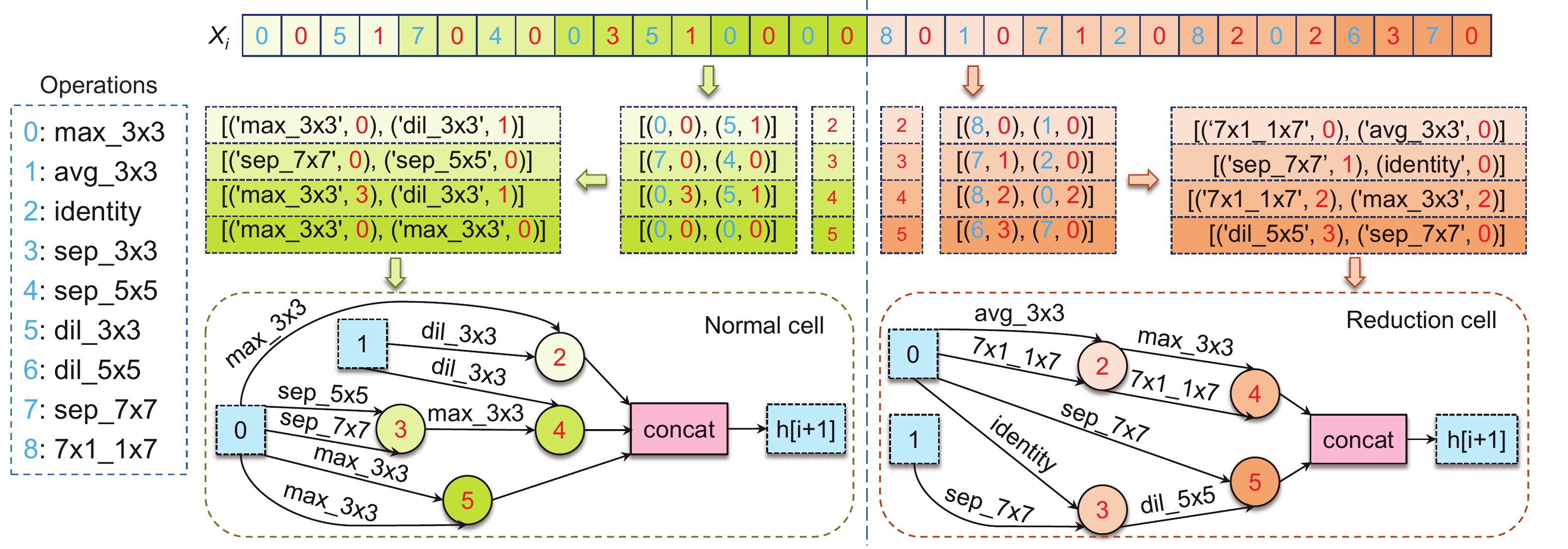}
\caption{An example of the encoding strategy}
\label{encode}
\end{figure*}

The encoding method adopted in this work is the same as the one in \cite{lu2019nsga} and a diagram of the encoding strategy is presented in Fig. \ref{encode}. Each node receives information from two other nodes. Take the node $B=4$ as an example, each individual is composed of $4\times2\times2\times2=32$ decision variables. As shown in Fig. \ref{encode}, $X_i$ represents an individual, the first 16 bits represent the genes in a normal cell, and the rest 16 bits represent the genes in a reduction cell.
The chromosome for each cell is divided into four segments, each consisting of four bits. The first and third bits in blue indicate which operations are to be performed, and second and fourth bits in red denote which nodes this node are connected to. 
Taking a normal cell as an example, which is coded by the first segment `[(0, 0), (5, 1)]', which means that the two inputs of node 2 are nodes 0 and 1, and operations `max\_3x3' and `dil\_3x3' are applied to nodes 0 and 1, respectively. The resulting features then added to be the output of the node. The nodes that are not connected to other nodes in the same cell, here nodes 2, 4, and 5, are then concatenated as the output of the entire cell.
\subsection{Objective Functions}\label{obj}
The objective functions are presented as following.
\begin{equation}
\begin{split}
\mathop {\min}: F(X)= \{ {f_1},{f_2}\} \\
%\mathop {\max} \{ {f_1},{f_2},{f_3}\} \\
\end{split}
\end{equation}
\subsubsection{Performance evaluation on Clean Examples}
In MORAS, we use the error rate as the measure to evaluate the performance of the architectures on a clean data without any attacks, which is $f_1$ of the multi-objective evolutionary algorithm:
\begin{equation}
\begin{split}
% \mathop {f_1} = J(\theta ,x,y)\\
 \mathop {f_1} = Err_{clean} = 1-(\frac{1}{N} \sum \mathbb{I}(\hat y==y))\times 100\%\\
\end{split}
\label{f1}
\end{equation}
where $Err_{clean}$ is the error rate on the original clean examples, $N$ is the number of examples, $y$ is the true label, and $\hat{y}$ is the predicted output. $\mathbb{I}$ is an indicator function, which outputs 1 if its input is true; otherwise, 0.
\subsubsection{Robustness against Adversarial Attacks}
Most existing work on search for robust neural architectures considers the robustness against one particular type of attacks. One reason is that it will be extremely time-consuming if we evaluate the performance of neural architectures under multiple types of adversarial attacks during the NAS. Neither is it very practical to the treat the performance of a model under one particular type of attacks as one separate objective. To computationally efficiently assess the robustness of neural architectures against multiple types of attacks during evolutionary NAS, one type of attacks will be randomly selected for each neural architecture so that over the generations, an evolving neural architecture will be assessed under various types of attacks. 

Note, however, that the ranges of the robustness values under various types of attacks may be very different. To reliably assess the overall robustness performance and to fairly compare the performance of different architectures that are trained using different types of adversarial samples within one generation, we normalize the robustness values using the mean and variance of the robustness values obtained on 18 hand-crafted neural networks before the evolutionary NAS starts. Thus, the robustness of a neural architecture under one type of adversarial attacks is calculated as follows, which is used as the second objective function in the evolutionary NAS:
\begin{eqnarray}
\label{f2a}
{f_{2}} &= &\frac{Err_{ad} -{\mu}_i}{{\sigma}_i}\\
\label{f2aa}
 Err_{ad} &= & 1-(\frac{1}{N} \sum \mathbb{I}(\hat y_a==y))\times 100
\end{eqnarray}
where $Err_{ad}$ the error rate on adversarial examples generated from a randomly selected type of adversarial attack, $\hat y_a$ is the predicted output, ${\mu}_i$ and ${\sigma}_i$ are the mean and the standard deviation of the error rate of different baseline architectures on the $i$-th adversarial attack.

\subsection{Overall Framework}

The two objectives previous defined are conflicting with each other and cannot be optimization simultaneously. Therefore, In this work, we employ a popular multi-objective evolutionary algorithm, namely the elitist non-dominated sorting genetic algorithm (NSGA-II) \cite{deb2002fast}, to search for architectures that trade off between accuracy and robustness.

\begin{algorithm}[htbp]\footnotesize{
\caption{The Framework of the proposed MORAS} \algblock{Begin}{End}
\label{alg1}
\textbf{Input:} The population size $N$, the maximal generation number $G$\\
\textbf{Output:} The non-dominated solutions in $\mathcal{S}$ after final adversarial training
\begin{algorithmic}[1]
\State \textbf{Initialization:} Set $\mathcal{S}$ = $\Phi$, generate $N$ parent individuals $P_0$ randomly with the gene encoding strategy, and evaluate each individual by using \textbf{the proposed evaluation approach}
\For {$t=0$ to $G$}
\State \textbf{Crossover and mutation:} Generate $N$ offspring $Q_t$ through SBX and PM
\State \textbf{Evaluation:} Calculate the fitness value $ f_1\ $ and $ f_2\ $ by using \textbf{the proposed evaluation approach}
\State \textbf{Merge:} $ {R_t} = {P_t} \cup {Q_t}\ $
\For {each individual in $ {R_t}\ $}
\State Do non-dominated sorting and calculate crowding distance
\State \textbf{Select} $N$ high-ranking solutions from $ {R_t} $
\EndFor
\State \textbf{Update} ${P_t}$
\State \textbf{Update} $\mathcal{S}$ with the individual in the global first non-dominated rank
\EndFor
\State \textbf{Final training:} Decode the non-dominated solutions from $\mathbb{P}$ for the final deep adversarial training;
\State \Return The trained architectures
\end{algorithmic}}
\end{algorithm}

As presented in Algorithm \ref{alg1}, MORAS starts with randomly generating an initial population $P_0$ of size $N$ with the given genetic encoding strategy (line 1) as described in Section \ref{3.1}. An archive $\mathcal{S}$, which is used to keep the global non-dominated solutions, is set to an empty set $\Phi$ initially.
%Each dimension of the individual is an integer and
All individuals in population $P_0$ are evaluated using the two objective functions, as detailed in Algorithm \ref{alg2}. The evolutionary process contains five steps (lines 3-9).
% which have been introduced in Section \ref{nsga2}.
Specifically, an offspring population is created by applying the simulated binary crossover (SBX) and polynomial mutation (PM) \cite{deb1995simulated} to the parents. This procedure repeats until $N$ offspring are generated. The offspring population ($Q_t$) is then combined with the parent population ($P_t$) to form a combined population $R_t$. The fast non-dominated sorting is then applied on $R_t$ to sort the individuals into a number of non-dominated fronts, in which the non-dominated solutions are ranked on the first front. A crowding distance is then calculated for solutions on the same front and then they are ranked in a descending order according to the crowding distance. Finally, the better half $R_t$ will be selected as the parent of the next generation $P_{t+1}$.  

Once the evolutionary search is completed, all non-dominated solutions in $\mathcal{S}$ are decoded to network architectures for complete deep training (line 13). This is because the training of individuals during the evolutionary process is often not sufficient due to a small number of epochs and a part of dataset for acceleration \cite{sun2018particle}.

\begin{algorithm}[!t]\footnotesize{
\caption{Performance Evaluation in MORAS} \algblock{Begin}{End}
\label{alg2}
\textbf{Input:}
The individual list $P_t$ of size $N$, training epoch $T$, batch size $M$, training set $\mathcal{D}_1$, validation set $\mathcal{D}_2$, the averaged error rate \textbf{$\mu$}=(${\mu}_1, {\mu}_2,\ldots,{\mu}_5$), the standard deviation \textbf{$\sigma$}=(${\sigma}_1, {\sigma}_2,\ldots, {\sigma}_5$)\\
\textbf{Output:}
The population $P_t$ with fitness $f_1$ and $f_2$
\begin{algorithmic}[1]
\For {$n=1$ to $N$}
\State \textbf{Decoding:} Decode the individual ${P_t^n}$ with decoding strategy to obtain a network model $M_n$
\For {$j=1$ to $T$}
\State //\textbf{Fast Adversarial training:} Train $M_n$ on $\mathcal{D}_1$ using FastAdv
\For{$i=1$ to $M$}
\State $\delta$ = Uiform(-$\epsilon$,$\epsilon$)
\State $\delta$ = $\delta$ + $\alpha$ $\cdot$ sign($\nabla_{\delta}l$($f_{\theta}(x_i+\delta), y_i$))
\State $\delta$ = max(min($\delta, \epsilon), -\epsilon$)
\State $\theta$ = $\theta$ - $\nabla_{\theta}l$($f_{\theta}(x_i+\delta), y_i$)
\EndFor
\EndFor
\State //\textbf{Validation:}
\State \textbf{$f_1^n$} $\leftarrow$ Test $M_n$ on $\mathcal{D}_2$ to get the $f_1^n$ according to Eq. (\ref{f1})
\State \textbf{$f_2^n$} $\leftarrow$ Randomly generate an integer $i$ from 0 to 5. Select the $i$th adversarial attack in $\lbrace$FGSM, BIM, PGD, FFGSM, FGSM-B$\rbrace$
%\begin{minipage}[t]{\linewidth}

Generate adversarial examples $\mathcal{A}$ by using the selected attack

Test $M_n$ on $\mathcal{A}$ to get the error rate according to Eq. (\ref{f2aa})

Use the corresponding ${\mu}_i$ and ${\sigma}_i$ to calculate $f_2^n$ as Eq. (\ref{f2a})

%\end{minipage}
\EndFor
\State \Return {$P_t$ with $f_1$ and $f_2$}
\end{algorithmic}}
\end{algorithm}

\begin{figure*}[!t]
\centering
\includegraphics[width=.9\textwidth]{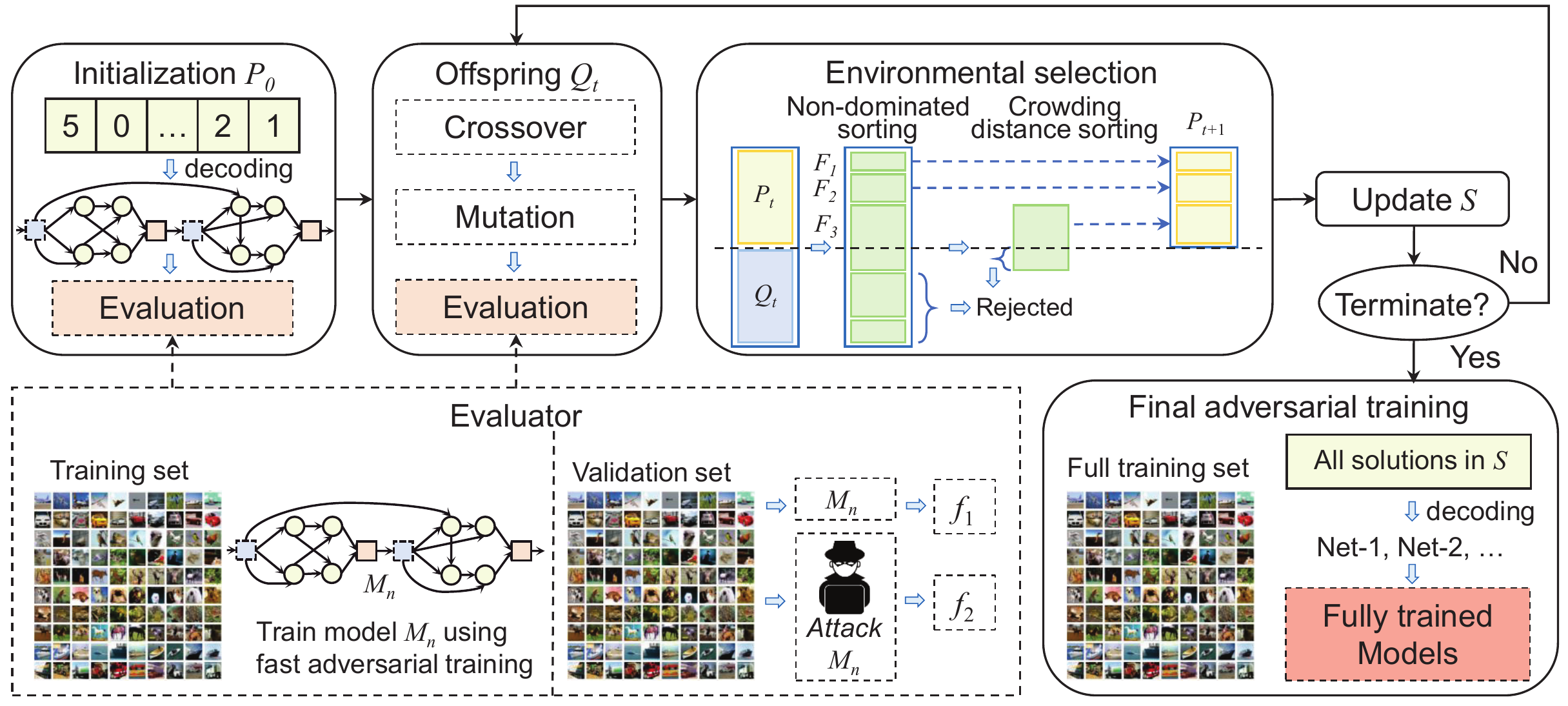}
\caption{A framework for multi-objective robust architecture search.}
\label{flow}
\end{figure*}

To further help understand the proposed algorithm, a diagram of MORAS is provided in Fig. \ref{flow}.

\section{Experiments}
To evaluate the performance of MORAS, we carry out experiments on two image classification tasks. In the following, we give a brief description of the used datasets, followed by an introduction of the manually designed CNNs as baselines for comparisons. Finally, the parameter settings of the proposed algorithm and the algorithms under comparison are presented. 
\subsection{Datasets}
We consider CIFAR-10 \cite{krizhevsky2010cifar} and CIFAR-100 \cite{krizhevsky2009learning} datasets as the classification tasks. CIFAR-10 and CIFAR-100 are labelled datasets which contain 10 and 100 classes, respectively. Both of them consist of a total number of 60,000 $ 32 \times 32$ pixel images. Therein, 50,000 images form the training set and the rest form the test set. 
Since it is time-consuming to train all the architectures on the full training set, we use only 24\% of the full training images during the architecture search process to improve the efficiency of the search. Specifically, during the search process, 10,000 images are used as the training set and 2000 as the validation set to estimate the performance. After evolution, all the architectures decoded from the global non-dominated solutions are re-trained using full 50,000 training images.

\subsection{Baseline Networks}\label{bsnets}
%\subsection{Training Standard Nets}
To estimate $\mu$ and $\sigma$ in Eq. (\ref{f2a}), we train five types of popular hand-crafted CNNs as the baselines. The architectures of the baselines and the 18 specific networks are as follows. VGG-11/13 \cite{simonyan2014very}, ResNet-18/34/50/101/152 \cite{he2016deep}, WideResNet-34, WideResNet-50-2/101-2 \cite{zagoruyko2016wide}, ResNeXt-50 \cite{xie2017aggregated}, PreAct ResNet-18 \cite{he2016identity}, DenseNet-121 \cite{huang2017densely}, MobileNet-V2 \cite{sandler2018mobilenetv2}, and ShuffleNet-V2-0.5$\times$/1$\times$/1.5$\times$/2$\times$ \cite{zhang2018shufflenet}.

\subsection{Parameter Settings}

The parameter settings of the experimental studies are divided into four parts as described below.

\begin{table}[]
\caption{Basic parameter configuration for various adversarial attacks}
\label{patk}
\centering
\begin{tabular}{cccc}
\hline
Attacks  & $\epsilon$ & $\alpha$ & Iteration \\ \hline
FGSM     & 8/255   & -     & -         \\
BIM      & 8/255   & 2/255 & 7         \\
% ILLC     & 4/255   & 1/255 & 5         \\
% RFGSM    & 16/255  & 8/255 & 1         \\
PGD      & 8/255     & 2/255 & 7         \\
FFGSM     & 8/255     & 12/255 & -         \\
Blk-FGSM & 0.007   & -     & -         \\ \hline
\end{tabular}
\end{table}
\subsubsection{Network architecture}
We follow the same convolutional search space defined by \cite{liu2018darts}. We set the number of phases $n_p$ to three and the number of nodes in each phase $n_o$ to seven. In most research, $n_o$ is set to four or five to reduce the computation burden. However, it is of significance to explore the robustness of architectures in a larger search space. The operation types are listed in Fig. \ref{encode}. We also fix the spatial resolution changes scheme similar to those in \cite{zoph2018learning}, in which a max-pooling with stride 2 is placed after the first and the second phases, and a global average pooling layer after the last phase. We set the number of filters (channels) in all node to 24 for each one of the generated network architecture.
\subsubsection{Adversarial attacks}

The parameter settings of different adversarial attacks, including FGSM, BIM, PGD, FFGSM, Blk-FGSM, are listed in Table \ref{patk}. We set the perturbation value $\epsilon$ and step size $\alpha$ as in their original paper. However, we set the attack iterations to a small number so that the accuracy will not be close to 0. We use the \textit{torchattacks} library provided by \cite{kim2020torchattacks} for all the adversarial attacks in our experiments.

\subsubsection{Evolutionary search}
Although we do not use the full dataset during searching, it is still highly time-consuming to evaluate an individual. So we use a relatively small population size 30 and the generation is set to 50. Hence, the total number of searched architectures in each compared algorithm is 1,500. The initial population is generated by uniform random sampling. The probabilities for crossover and mutation are set to 0.9 and 0.02, respectively.

\subsubsection{Training process}\label{train}
During architecture search, we train each architecture on a subset of the training data for 15 epochs. The batch size is 128. All architectures are trained using FastAdv \cite{wong2020fast} and a cyclic learning rate \cite{smith2019super}, which schedules the learning rate linearly from zero to a predefined maximum learning rate and back down to zero. Using a cyclic learning rate allows the architectures to converge to the benchmark accuracies in tens of epochs instead of hundreds. The maximum learning rates are set to 0.2 and 0.05 for CIFAR-10 and CIFAR-100, respectively. After obtaining the non-dominated solutions, we decode them and train the obtained networks on the whole training set for 30 epochs. 

\subsection{Compared algorithms}
To valid the performance of the MORAS, we compare the proposed algorithm with the following two variants:
\begin{itemize}
\item Instead of evaluating the robustness on multiple types of attacks, the robustness on adversarial examples generated from one white-box attack only, FGSM is maximized together with the performance on clean data. All other settings are the same as the proposed algorithm. The neural architectures decoded from the obtained non-dominated solutions are named MORNet-F.
\item A single-objective genetic algorithm for neural architecture search to maximize the performance on clean images. The obtained best architecture is named SGANet.
\end{itemize}
For fair comparisons, we train all network architectures found by the compared algorithms in the same way as described in Subsection \ref{train}.

% \section{Experimental Results and Discussions}
\section{Experimental Results}
\begin{figure}[t]
\centering
\subfigure[The HV values of MORAS over the generations on CIFAR-10.]{
\includegraphics[width=.35\textwidth]{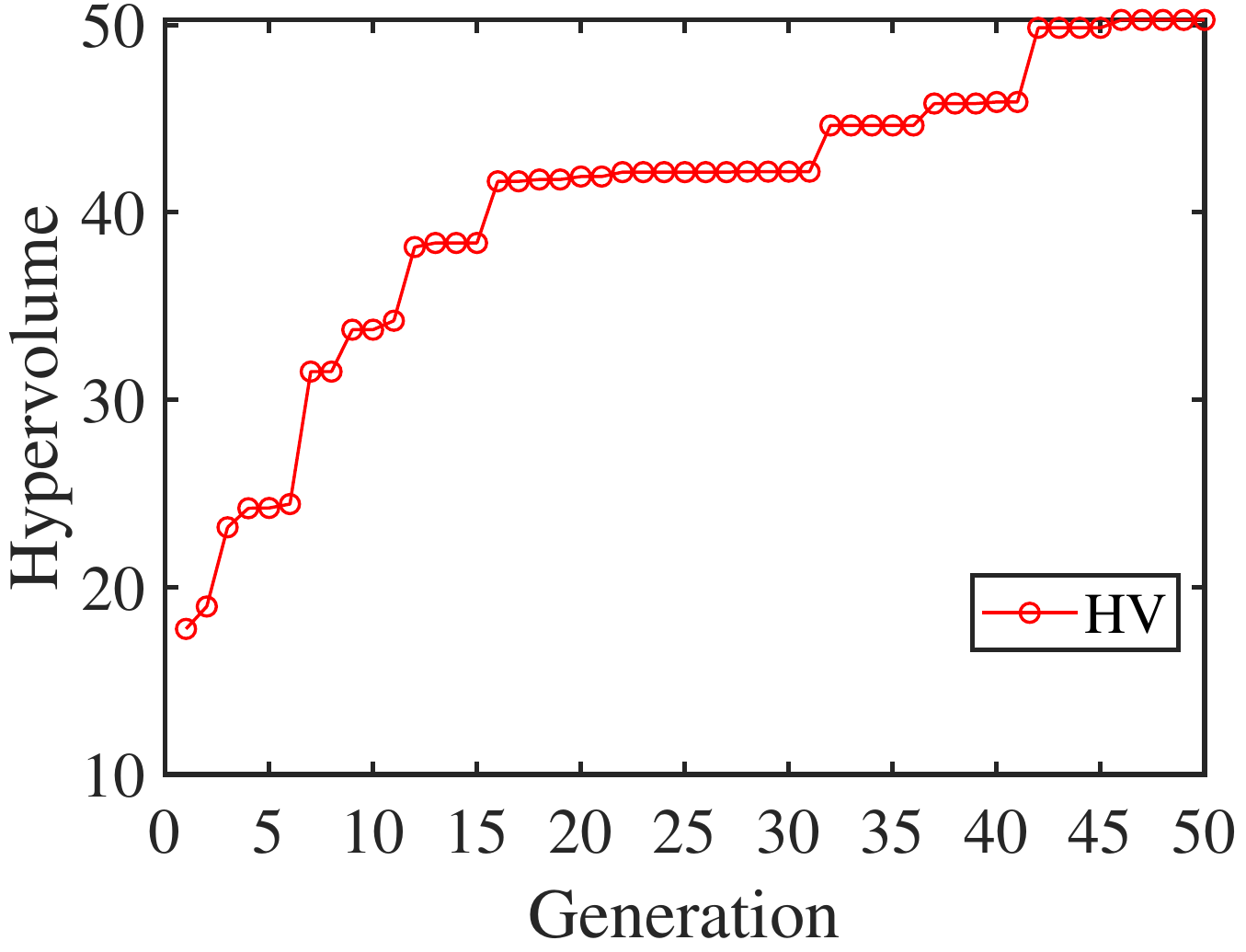}
\label{HV1}
%\caption{fig1}
}
\quad
\subfigure[Non-dominated solutions found by MORAS.]{
\includegraphics[width=.35\textwidth]{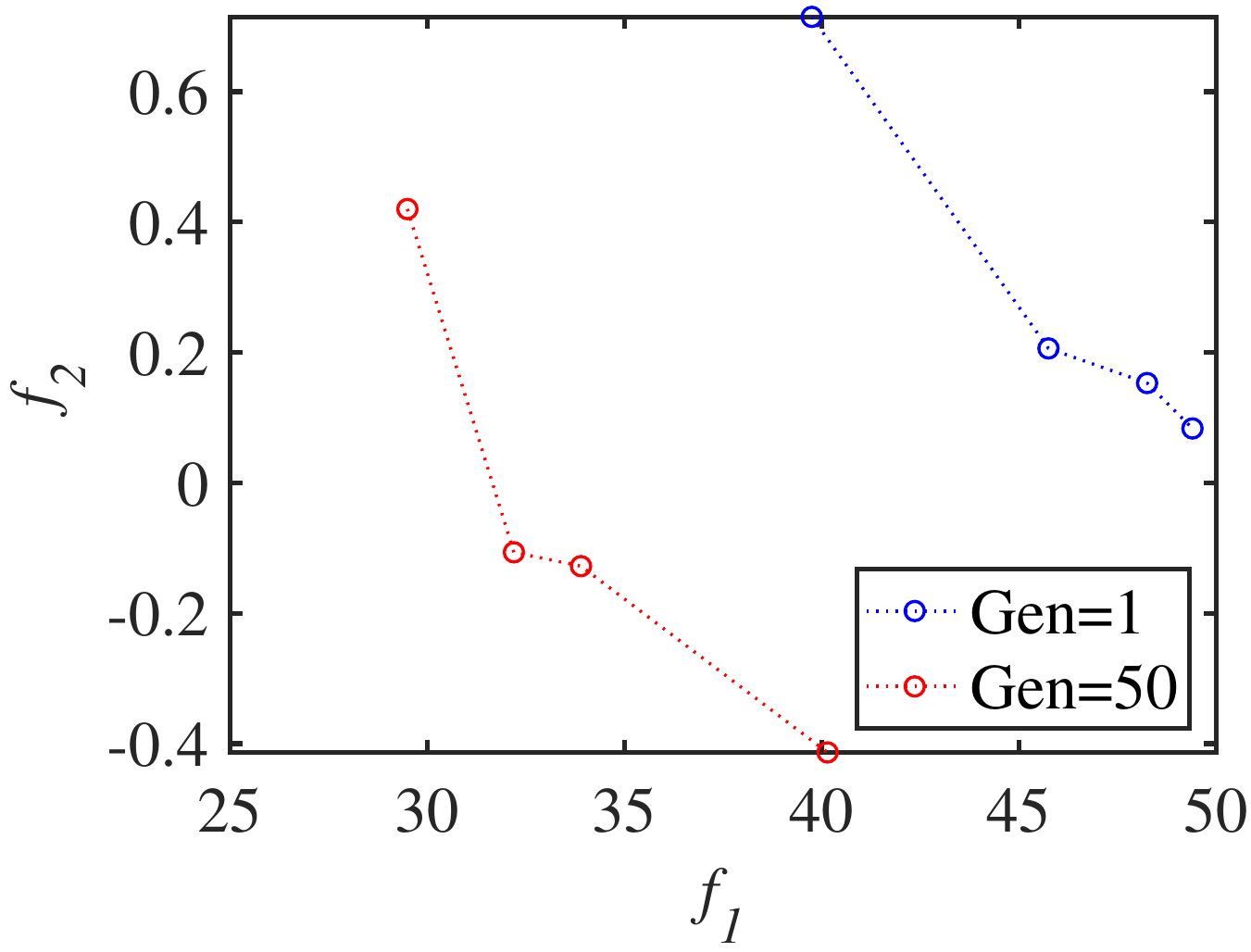}
\label{PF1}}
\caption{The results of MORAS on CIFAR-10.}
\label{E1}
\end{figure}

We present the evaluation results on CIFAR-10 in Section \ref{c10}, and CIFAR-100 in Section \ref{c100}. In all experiments, each individual is trained for 15 epochs on a subset of the training data. Thereafter, individuals are evaluated based on the objective functions. The non-dominated solutions are stored in an external archive. The non-dominated solutions obtained after 50 generations present a set of trade-off solutions between accuracy and robustness. The performance of the architectures obtained by the proposed approach are then compared to those obtained in Section \ref{AE}. 
% Finally, we discuss the experimental results in Section \ref{AD}.

Hypervolume (HV) \cite{while2006faster} is used as the performance indicator to assess the convergence property of the multi-objective search algorithms. The larger the HV value is, the better the performance. The reference point for calculating the hypervolume is set to $(1.1\times{max(f_1)},1.1\times{max(f_2)})$ in all experiments.

\subsection{Performance of MORNet on CIFAR-10}\label{c10}

The search results of the proposed MORAS are presented in Fig. \ref{E1}. From Fig. \ref{HV1}, it can be observed that hypervolume values increase during the evolutionary process. Fig. \ref{PF1} shows the non-dominated solutions in the objective space obtained in the first and last generation, which are denoted by blue circles and red circles, respectively. We sort the four non-dominated solutions in the final generation in an ascending order according to their $f_2$ values and denote them as MORNet-V1-1, MORNet-V1-2, MORNet-V1-3, MORNet-V1-4.

\begin{table*}[htbp]
\caption{Standard and robust performance of the searched solutions on CIFAR-10. We compare the non-dominated solutions obtained by the competitors. All models are adversarially trained using FastAdv for 30 epochs. The training hyper-parameters are the same for all models. The parameters of adversarial attacks are set the same as Table \ref{patk}. For all items, larger is better; best results are highlighted in bold.}
\label{Res10}
\centering
\resizebox{\textwidth}{!}{
\begin{tabular}{cccccccc}
\hline
Networks    & Clean (\%) & FGSM (\%)  & BIM (\%)   & PGD (\%)   & FFGSM (\%) & Blk-FGSM (\%) & Robustness \\ \hline
MORNet-V1-1 & 83.86 & 40.24 & 73.79 & 37.72 & 38.03 & 42.80    & -0.11      \\
MORNet-V1-2 & 83.29 & 38.55 & 73.56 & 36.21 & 36.59 & 40.93    & -20.42     \\
MORNet-V1-3 & 84.10 & \textbf{42.55} & 74.40 & \textbf{39.71} & \textbf{40.11} & \textbf{45.02}    & \textbf{27.73}      \\
MORNet-V1-4 & \textbf{84.24} & 41.00 & \textbf{74.77} & 38.28 & 38.70 & 43.35    & 10.69      \\
MORNet-F1-1 & 81.52 & 40.48 & 72.19 & 37.89 & 38.22 & 43.18    & -2.29      \\
MORNet-F1-2 & 81.96 & 39.73 & 72.14 & 37.33 & 37.61 & 42.50    & -10.28     \\
MORNet-F1-3 & 83.32 & 40.72 & 73.68 & 37.90 & 38.26 & 43.38    & 3.95       \\
SGANet      & 83.58 & 39.47 & 73.58 & 36.97 & 37.39 & 42.14    & -9.27      \\ \hline
\end{tabular}}
\end{table*}

After the multi-objective robust architecture search process, we train the searched MORNets using FastAdv \cite{wong2020fast} for 30 epochs on the full training dataset. To show the advantages of the proposed MORAS over the compared approaches, we present their test results on the clean test set and different adversarial attacks in Table \ref{Res10}. Specifically, Table \ref{Res10} is divided into eight columns: the first column shows the names of the compared algorithms; the second column shows the classification accuracy (\%) of different architectures on the clean test set, the third to sixth columns denote the classification accuracy on white-box adversarial attacks (FGSM, BIM, PGD, and FFGSM), the performance of different models on the black-box adversarial attack is listed in the seventh column. In addition, the last column shows the robustness of the models against the above-mentioned attacks. The best results are highlighted in bold.

As can be clearly seen from the second column in Table \ref{Res10}, on the clean test set,  MORNet-V1-4 achieves 84.24\% accuracy and performs best among all the architectures. MORNet-V1-3 and MORNet-V1-3, which achieve 84.1\% and 83.86\%, respectively, still outperform MORNets and SGANet on the original test set. 

Under adversarial attacks, MORNet-V1-3 outperforms other models under comparison on FGSM, PGD, FFGSM, and Blk-FGSM. MORNet-V1-4 achieve the best results against BIM and second-best results against FGSM, PGD and FFGSM. 

\begin{figure}[t]
\centering
\subfigure[Normal Cell]{
\includegraphics[width=.4\textwidth]{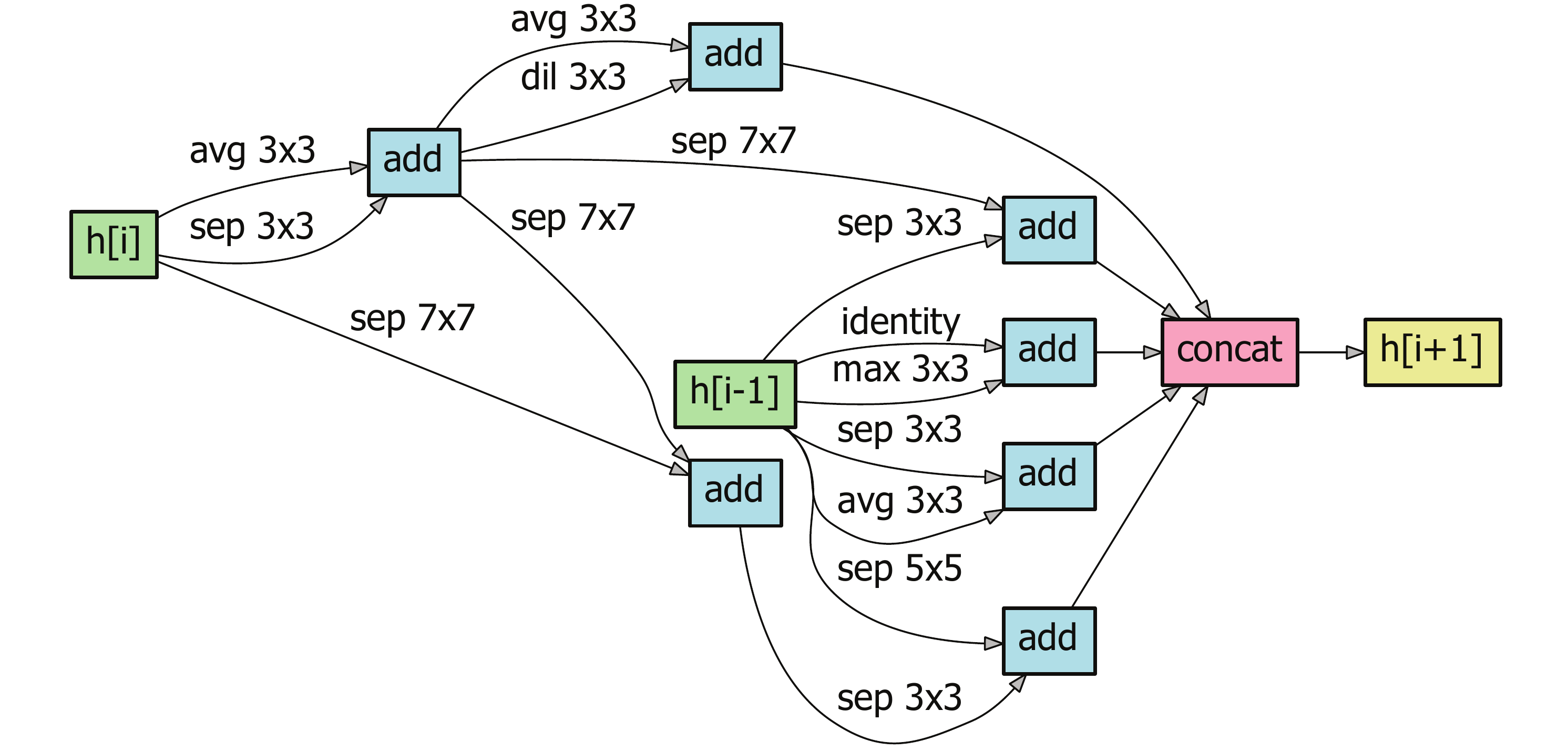}
\label{v10_n}
}
\quad
\subfigure[Reduction Cell]{
\includegraphics[width=.4\textwidth]{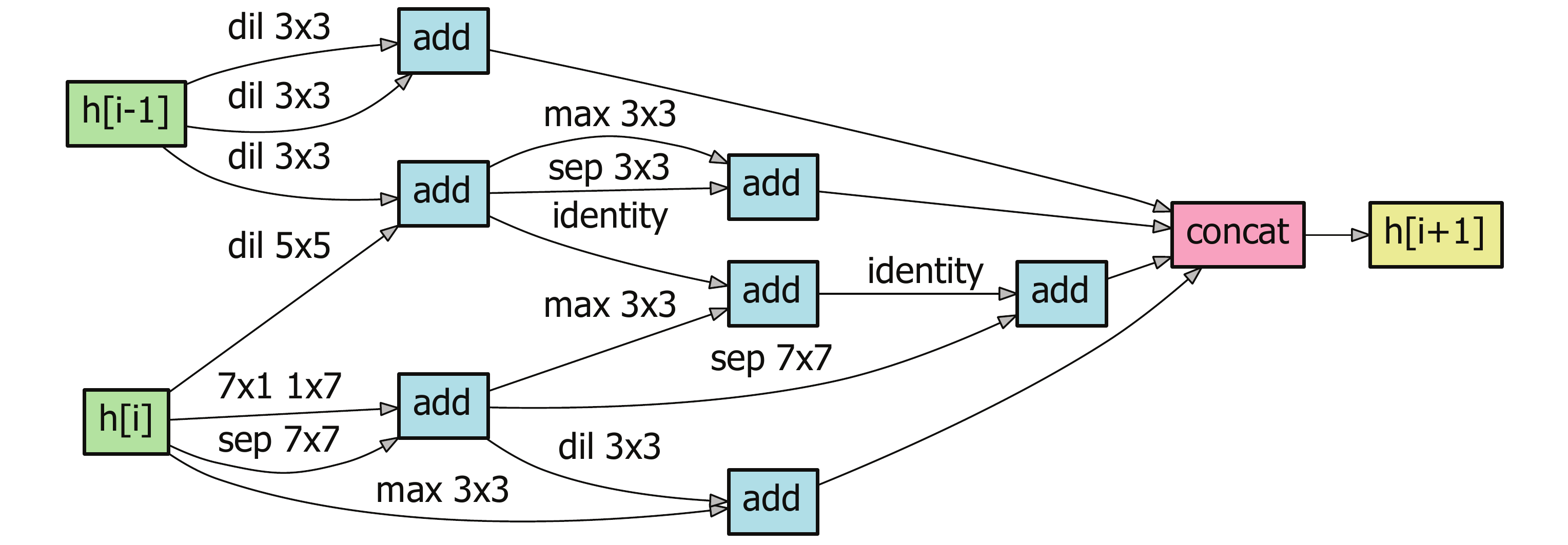}
\label{v10_r}}
\caption{Normal and reduction convolutional cell architectures found by MORAS on CIFAR-10 dataset.}
\label{v10}
\end{figure}

To take a closer look, we calculate the robustness of each architecture according to the method proposed in \cite{chang2020evaluating} so that we can clearly compare the performance of various architectures on different adversarial attacks. The robustness is measured based on the classification accuracy under five adversarial attacks. From the results presented in the last column in Table \ref{Res10}, we can see that the robust values of the proposed MORNet-V1-3 and MORNet-V1-4 are higher than all competitors. Therefore, these result demonstrate the effectiveness of the proposed approach. The obtained normal cell and reduction cell of MORNet-V1-4 architectures are visualized in Fig. \ref{v10}. 

\subsection{Performance of MORNet on CIFAR-100}\label{c100}
\begin{figure*}[t]
\centering
\subfigure[The HV values of MORAS over the generations on CIFAR-100.]{
\includegraphics[width=.34\textwidth]{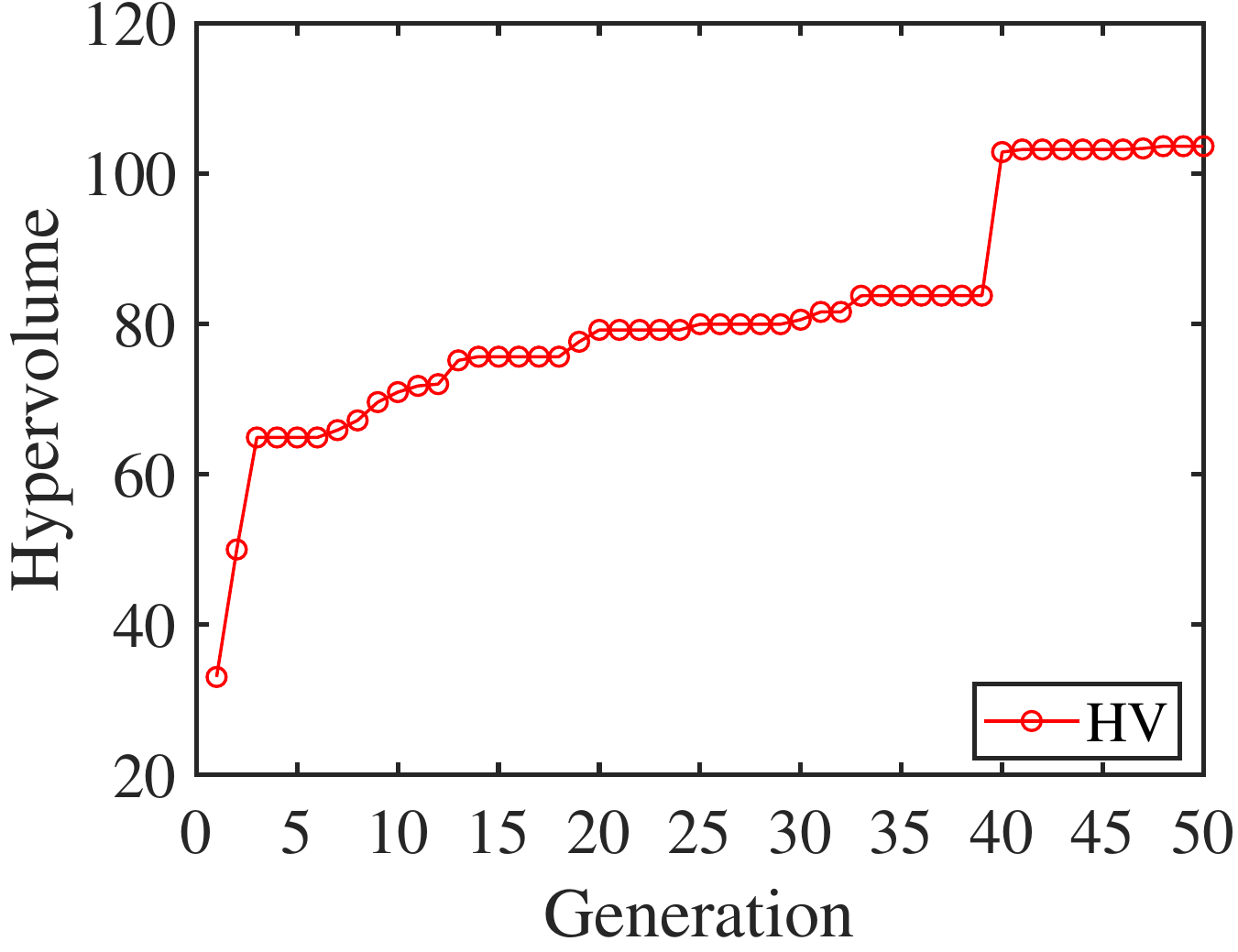}
\label{HV2}
}
\quad
\subfigure[Non-dominated solutions obtained by MORAS.]{
\includegraphics[width=.35\textwidth]{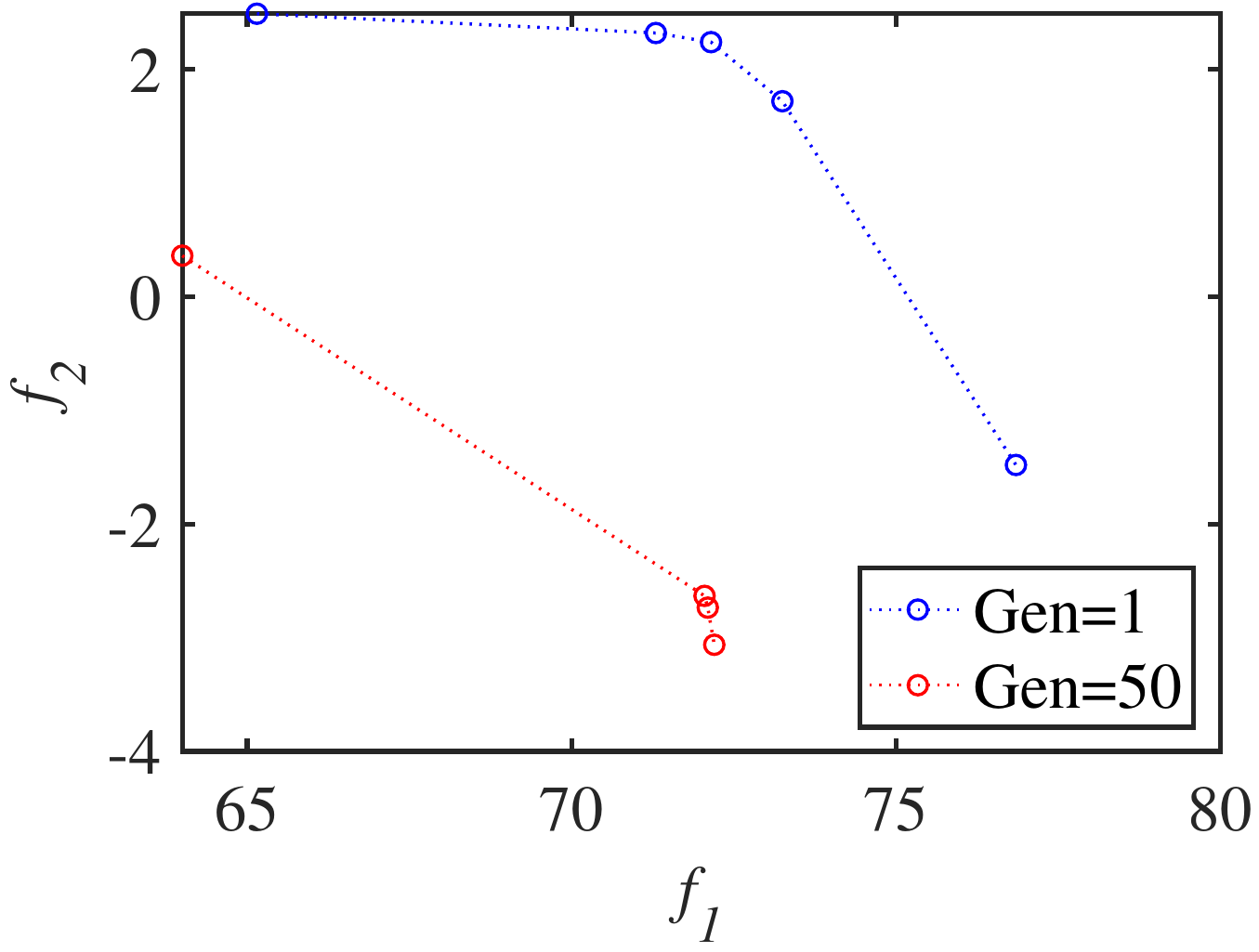}
\label{PF2}
}
\caption{The results of MORAS on CIFAR-100}
\label{E2}
\end{figure*}

We show the search results of the proposed MORAS on CIFAR-100 in Fig. \ref{E2}. From Fig. \ref{HV2}, it can be observed that hypervolume values increase during evolution. Fig. \ref{PF2} shows the non-dominated solutions in the objective space obtained in the first and last generations. The four non-dominated solutions obtained in the final generation, which are sorted in an ascending order according to $f_2$ values, are referred to as MORNet-V2-1, MORNet-V2-2, MORNet-V2-3, MORNet-V2-4.

\begin{table*}[htbp]
\caption{Standard and robust performance of the searched solutions on CIFAR-100. We compare the non-dominated solutions obtained by the compared algorithms. All models are adversarially trained using FastAdv for 30 epochs. The training hyper-parameters are the same for all models. All parameters of adversarial attacks are set the same as those in Table \ref{patk}. For all items, larger is better; the best results are highlighted in bold.}
\label{Res100}
\centering
\resizebox{\textwidth}{!}{
\begin{tabular}{ccccccccccc}
\hline
Networks & Clean (\%) & FGSM (\%)  & BIM (\%)   & PGD (\%)   & FFGSM (\%) & Blk-FGSM (\%) & Robustness \\ \hline
MORNet-V2-1       & 59.84 & 18.28 & 44.17 & 16.09 & 16.46 & 19.98    & 7.71       \\
MORNet-V2-2       & 58.83 & 18.00 & 43.45 & 16.13 & 16.31 & 19.74    & 4.51      \\
MORNet-V2-3       & 59.82 & 17.66 & 43.86 & 15.49 & 15.80 & 19.38    & 1.09       \\
MORNet-V2-4       & \textbf{60.02} & \textbf{18.92} & \textbf{44.52} & \textbf{16.40} & \textbf{16.78} & \textbf{20.85}    & \textbf{13.62} \\
MORNet-F2-1       & 56.77 & 16.90 & 42.63 & 15.05 & 15.31 & 18.48    & -7.98      \\
MORNet-F2-2       & 57.34 & 16.71 & 42.21 & 14.69 & 15.04 & 18.13    & -11.76     \\
SGANet            & 59.66 & 17.05 & 43.49 & 14.67 & 14.93 & 18.57    & -7.17      \\ \hline
\end{tabular}}
\end{table*}

The training procedure is similar to that on CIFAR-10. Table \ref{Res100} shows the performance on the clean test set and different adversarial attacks. The best results are highlighted in bold.

It can be observed that on the clean test set,  MORNet-V2-4 achieves 60.02\% accuracy and performs the best among all the architectures. MORNet-V2-1 and MORNet-V2-3 also outperform MORNet-F2-1, MORNet-F2-2 and SGANet with regards to the accuracy on the clean data. 

\begin{figure}[t]
\centering
\subfigure[Normal Cell]{
\includegraphics[width=.5\textwidth]{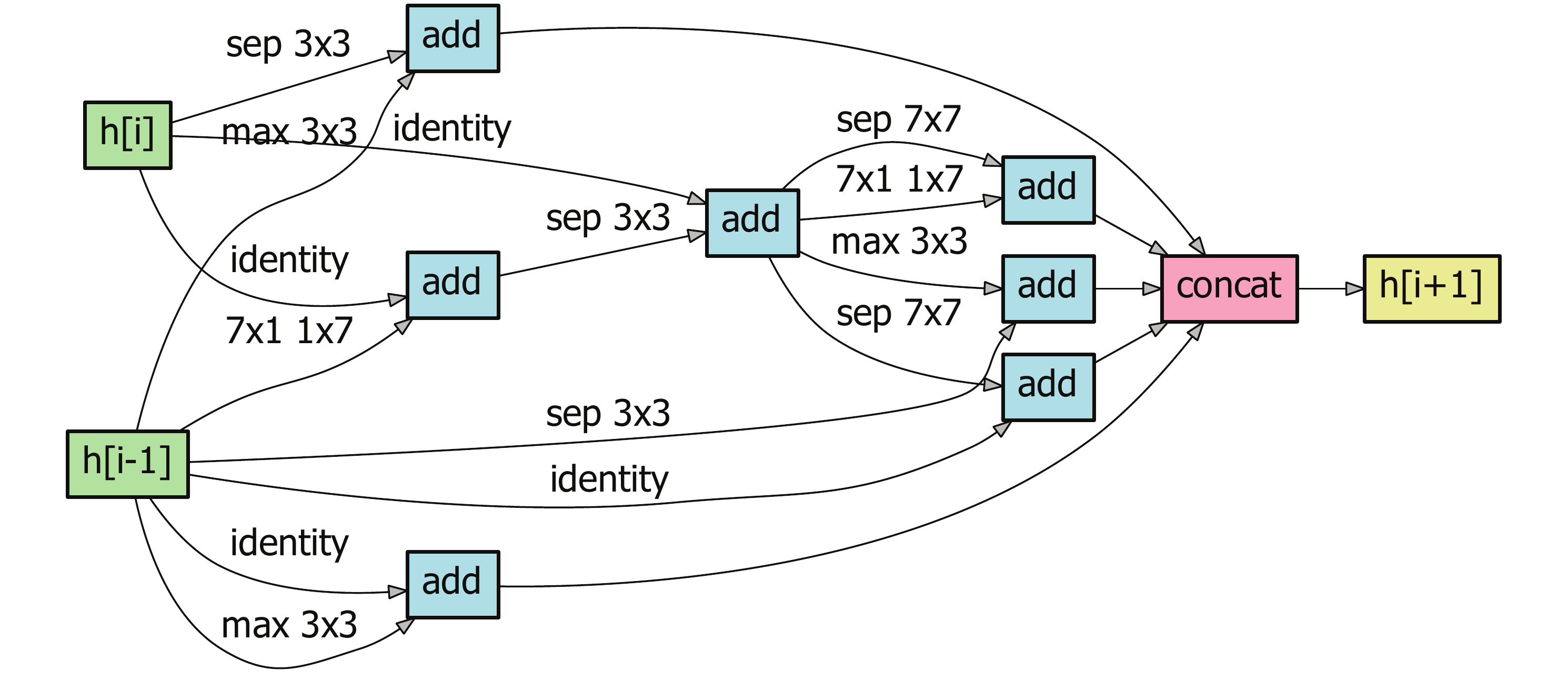}
\label{v100_n}
}
\quad
\subfigure[Reduction Cell]{
\includegraphics[width=.4\textwidth]{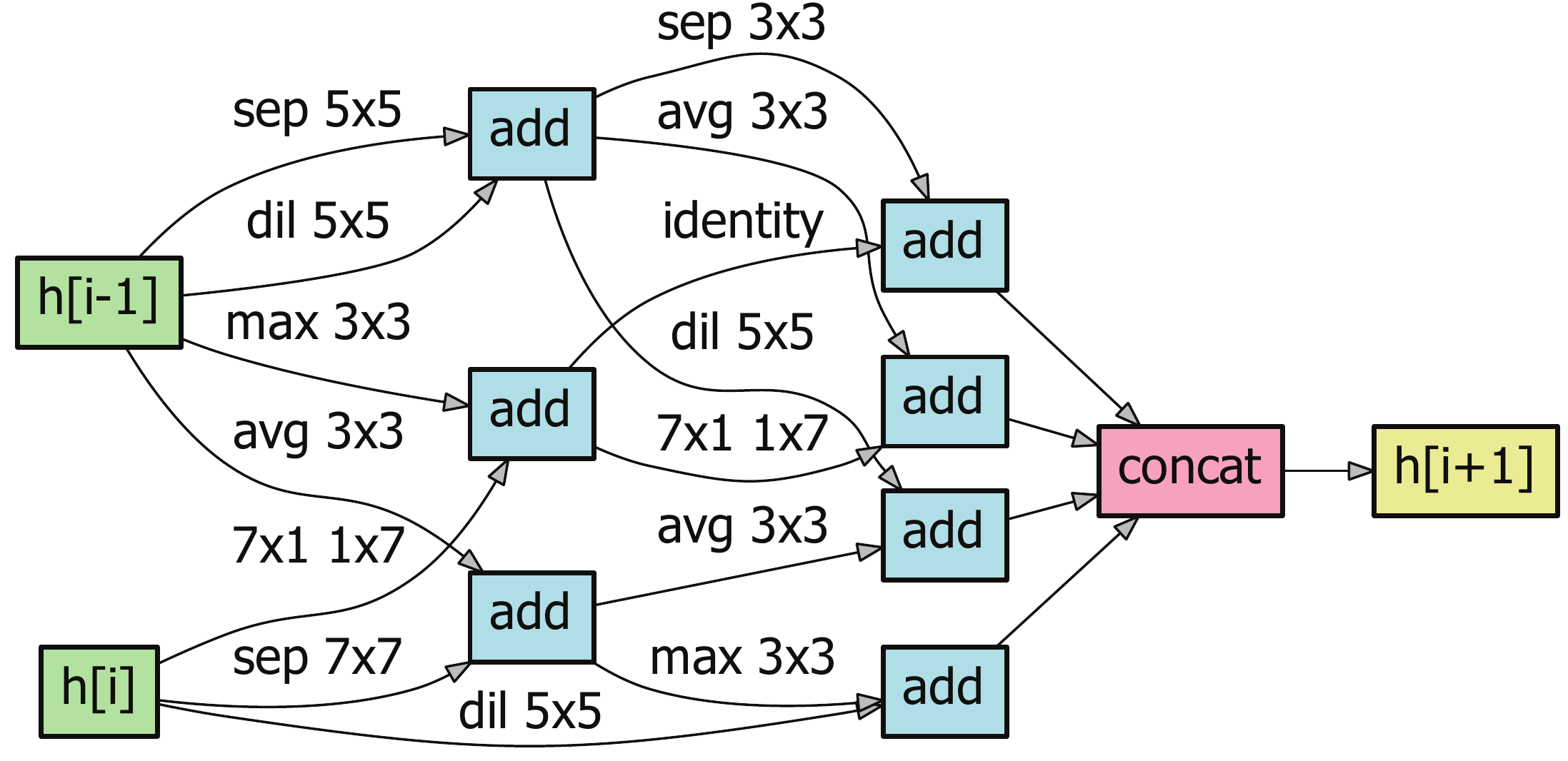}
\label{v100_r}}
\caption{Normal and reduction convolutional cell architectures found by MORAS on CIFAR-100 dataset.}
\label{v100}
\end{figure}

Remarkably, our MORNet-V2-4 achieves highest robust accuracy among peer competitors on all adversarial attacks. From the results in the last column in Table \ref{Res100}, we can see that the robustness values of MORNet-V2-1, MORNet-V2-2, MORNet-V2-3 and MORNet-V2-4 are positive and larger than that of MORNet-F2-1, MORNet-F2-2 and SGANet, indicating that the proposed MORAS is effective in search for robust architectures. The normal cell and reduction cell of MORNet-V2-4 architectures are visualized in Fig. \ref{v100}.

\subsection{Comparison with existing robust architectures}\label{AE}
\begin{figure}[t]
\centering
\subfigure[CIFAR-10]{
\includegraphics[width=.4\textwidth]{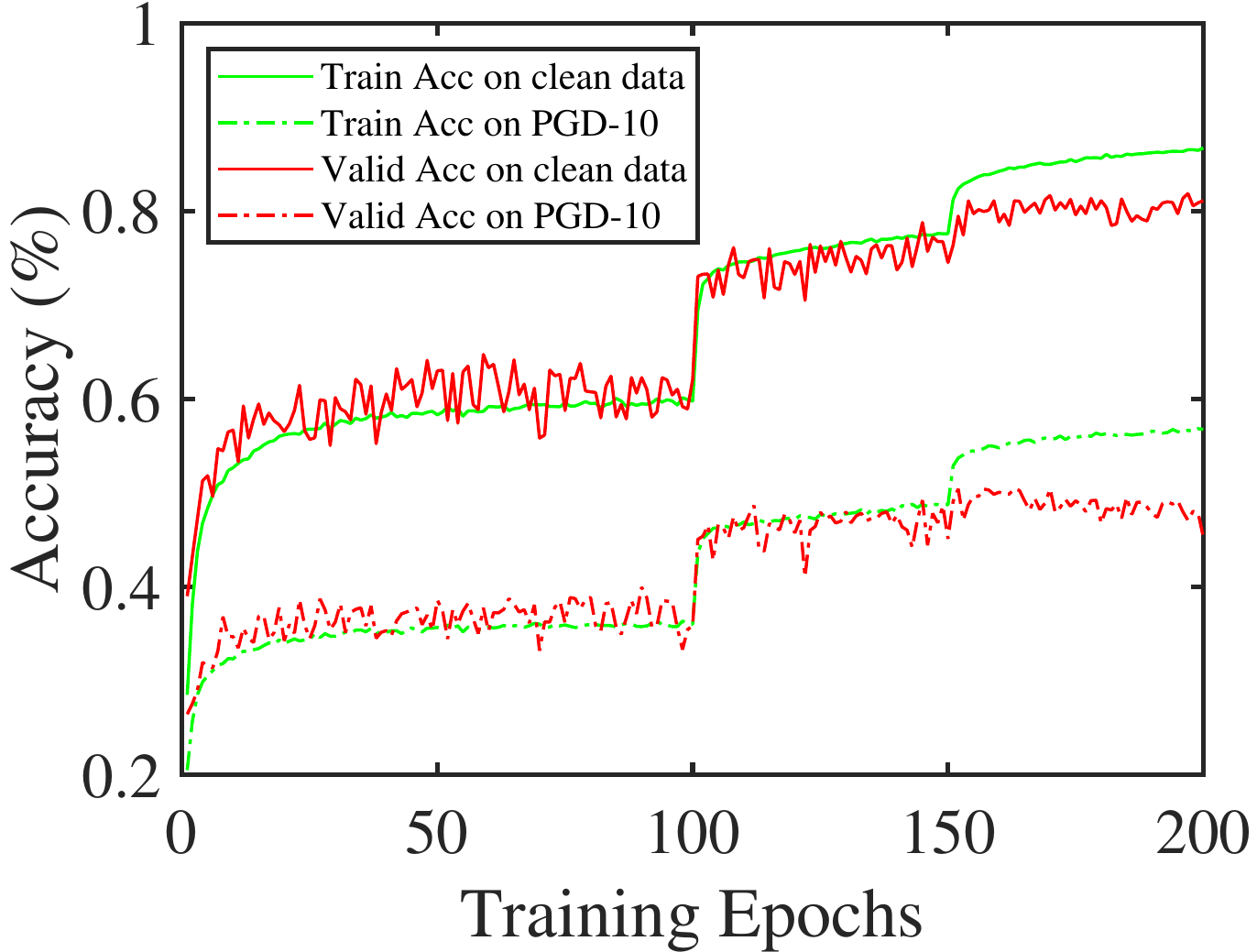}
\label{c10_200}
}
\quad
\subfigure[CIFAR-100]{
\includegraphics[width=.4\textwidth]{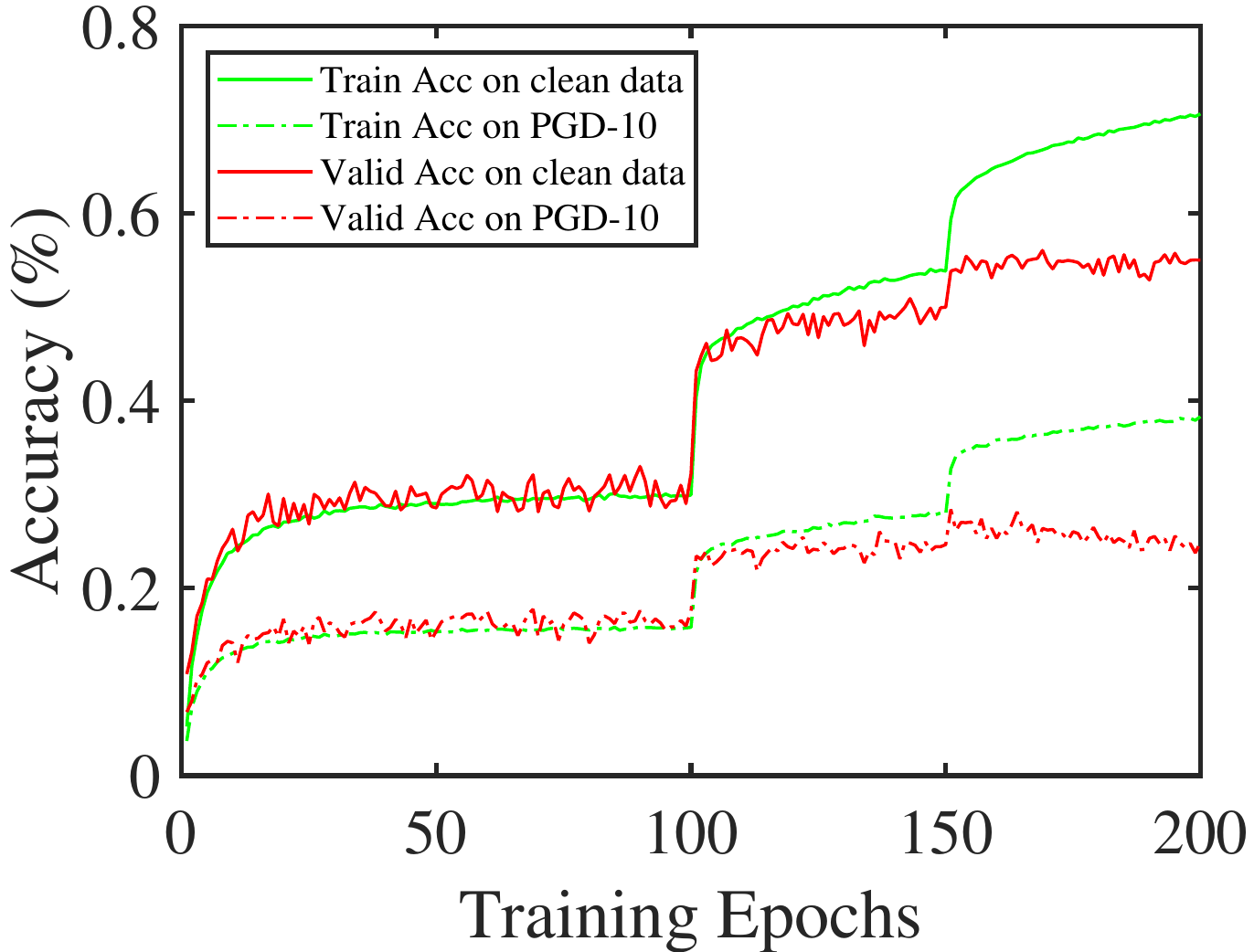}
\label{c101_200}}
\caption{Learning curves for FastAdv+ adversarial training plotting the accuracy.}
\label{LC}
\end{figure}

In the above experiments, we use FastAdv \cite{wong2020fast} to train all the architectures with a small number of epochs to reduce the evaluation cost. In this section, we compare the architectures having the highest robustness value in the proposed MORAS family with the following networks: the widely used PreAct ResNet-18 \cite{he2016identity} and WideResNet-34 \cite{zagoruyko2016wide} in adversarial literature; RobNet \cite{guo2020meets}, which is obtained from one-shot NAS with PGD-AT. We use FastAdv+ \cite{li2020towards} to train the models for 200 epochs and the learning curves on CIFAR-10 and CIFAR-100 are shown in Fig. \ref{LC}. Table \ref{ComRes} shows the performance of and the robust architectures obtained by MORAS on the CIFAR-10 and CIFAR-100 datasets. Note that the results of RobNet in Table \ref{ComRes} are extracted from \cite{guo2020meets}. 

In addition to the four white-box adversarial attacks and a transferable black-box adversarial attack, we also test these architectures on PGD-10/20/50, which are the PGD attacker with 10, 20 and 50 attack iterations. Although PGD with a larger iteration number means a stronger adversary, the performance of the above models remains stable as the number of iteration increases.

As shown in Table \ref{ComRes}, the performance of the architectures obtained by MORAS outperforms the compared existing robust networks under the attack of FGSM, PGD-7/10/20/50, FFGSM and Blk-FGSM on both CIFAR-10 and CIFAR-100 datasets. Although our architectures did not achieve higher accuracy than PreAct ResNet-18 on the clean dataset, they produce satisfying accuracies compared to Wide ResNet-34 and RobNet-free.
% \usepackage{multirow}

\begin{table*}[htbp]
\caption{Performance of networks on CIFAR-10 and CIFAR-100. We compare the best MORNet models with the state-of-the-art architectures.  The results of RobNets are extracted from \cite{guo2020meets}, and other models are adversarially trained using FastAdv+ \cite{li2020towards}.}
\label{ComRes}
\centering
\resizebox{\textwidth}{!}{
\begin{tabular}{ccccccccccc}
\hline
Datasets                   & Models           & Clean (\%) & FGSM (\%) & BIM (\%) & PGD-7 (\%) & PGD-10 (\%) & PGD-20 (\%) & PGD-50 (\%) & FFGSM (\%) & Blk-FGSM (\%) \\ \hline
\multirow{4}{*}{CIFAR-10}  & PreAct ResNet-18 & 83.54          & 55.50     & \textbf{67.89}    & 48.47      & 48.45       & 48.43       & 48.43       & 48.47      & 59.05         \\
                           & WideResNet-34    & \textbf{86.52}          & 53.57     & 67.65    & 47.10      & 47.10       & 47.10       & 46.90       & 47.10      & 56.75         \\
                           & RobNet-free      & 82.79          & 58.38     & -        & -          & -           & 52.74       & 52.57       & -          & 65.06         \\
                           & \textbf{Ours}             & 82.82          & \textbf{59.42}     & 66.18    & \textbf{58.56}      & \textbf{58.44}       & \textbf{58.42}       & \textbf{58.41}       &\textbf{ 58.87}      & \textbf{66.20}          \\ \hline
\multirow{3}{*}{CIFAR-100} & PreAct ResNet-18 & \textbf{60.78}          & 30.35     & \textbf{47.51}    & 28.63      & 28.04       & 28.02       & 28.01       & 28.30      & 31.50         \\
                           & WideResNet-34    & 60.57          & 30.84     & 44.93    & 29.53      & 29.11       & 28.61       & 28.61       & 29.34      & 32.94         \\
                           & \textbf{Ours}             & 59.98          & \textbf{35.72}     & 42.24    & \textbf{35.02 }     & \textbf{34.55}       & \textbf{34.56}       & \textbf{34.56}       & \textbf{35.11}      & \textbf{41.59}         \\ \hline
\end{tabular}}
\end{table*}
% \subsection{Discussion}\label{AD}
% Since partial training is used in the evolutionary search for reducing the computation time, the non-dominace relationship may change once the non-dominated solutions are fully trained. For example, MORNet-V1-1 and MORNet-V2-1 show the highest robustness values when the evolutionary search is complete. However, MORNet-V1-3 and MORNet-V2-4 show better robustness after full training on CIFAR-10 and CIFAR-100, respectively. This, however, may happen to all NAS methods that use any proxy evaluation methods, and the proposed MORAS is still effective in finding neural architectures that are robust to multiple types of adversarial attacks.  

% When conducting the experiments comparing with the state-of-the-art architectures, we choose an adversarial training approach to train the MORNets which are both effective and efficient. It should be noted that most researches developed adversarial training techniques based on specific networks, such as PreAct ResNet and WideResNet. Besides, most existing research on adversarial robustness uses the CIFAR-10 dataset for empirical validation and not much work has been reported on CIFAR-100. Hence, more investigations are needed for tuning the hyper-parameters in training the existing networks on CIFAR-100. 

\section{Conclusions and Future Work}
Deep neural networks are vulnerable to adversarial examples, which becomes one of the most important research topics in deep learning. While most existing work considers the robustness of neural networks on one single type of attacks, this paper proposes a multi-objective evolutionary approach to discover neural architectures that are relatively insensitive to multiple types of adversarial attacks. To reduce the computing budget during the search, we randomly select one type of attacks from four widely used white-box adversarial attacks and a transferable black-box adversarial attack, instead of averaging over all types of attacks. Our results demonstrate that the proposed algorithm achieves the best overall performance against different types of attacks among all compared methods on the CIFAR-10 and CIFAR-100 datasets, confirming the effectiveness of the proposed algorithm. 

A few limitations remain to be addressed. For example, it is still very time-consuming to train all the offspring architectures to obtain their fitness values during the evolutionary search process. Thus, our future work will investigate computationally more efficient methods for evaluation of the robustness against multiple attacks. Besides, this work uses the same operations as those in DARTS, which may limit the overall robustness of the searched architectures. It is of interest to explore more effective operations to defend against various types of adversarial attacks. Finally, it is worth exploring more efficient and effective adversarial training methods for various NAS search spaces.

% \section{Acknowledgements}

% \appendix
% \section{Appendix 1}
% Here, we give the definitions of different objective functions:

\printcredits

\bibliographystyle{model1-num-names}

\bibliography{my_refs}


%Author biography without author photo.
%Author biography. Author biography. Author biography.
%Author biography. Author biography. Author biography.
%Author biography. Author biography. Author biography.
%Author biography. Author biography. Author biography.
%Author biography. Author biography. Author biography.
%Author biography. Author biography. Author biography.
%Author biography. Author biography. Author biography.
%Author biography. Author biography. Author biography.
%Author biography. Author biography. Author biography.
%\bio{figs/pic1}
%Author biography with author photo.
%Author biography. Author biography. Author biography.
%Author biography. Author biography. Author biography.
%Author biography. Author biography. Author biography.
%Author biography. Author biography. Author biography.
%Author biography. Author biography. Author biography.
%Author biography. Author biography. Author biography.
%Author biography. Author biography. Author biography.
%Author biography. Author biography. Author biography.
%Author biography. Author biography. Author biography.
%\bio{figs/pic1}
%Author biography with author photo.
%Author biography. Author biography. Author biography.
%Author biography. Author biography. Author biography.
%Author biography. Author biography. Author biography.
%Author biography. Author biography. Author biography.

\end{document}